\documentclass[11pt]{article}

\usepackage{hyperref}
\usepackage[T1]{fontenc}
\usepackage{graphicx}
\usepackage[dvipsnames]{xcolor} 
\usepackage{tcolorbox} 
\usepackage{amssymb}
\usepackage{listings}
\usepackage{bbding}
\definecolor{mybgcolor}{rgb}{0.95, 0.95, 0.92}
\usepackage{amsmath,amsfonts}
\usepackage{algorithmic}
\usepackage{graphicx}
\usepackage{textcomp}
\usepackage{xcolor}
\usepackage{array}
\usepackage{makecell}
\usepackage{listings}
\usepackage{multirow}
\usepackage{amsmath}
\usepackage{float}
\usepackage{dblfloatfix}   
\usepackage{placeins}      
\usepackage{makecell}
\usepackage{caption}

\newtcolorbox{promptbox}[1][]{
  colback=yellow!5,      
  left=2mm,               
  right=2mm,              
  boxrule=0.5pt,          
  arc=2mm,                
  auto outer arc,
  fontupper=\ttfamily\small, 
  colbacktitle=yellow!20,  
  coltitle=black, 
  #1
}

\newfloat{listing}{htbp}{lop}
\floatname{listing}{Listing}

\def\BibTeX{{\rm B\kern-.05em{\sc i\kern-.025em b}\kern-.08em
    T\kern-.1667em\lower.7ex\hbox{E}\kern-.125emX}}

\begin{document}

\title{On the use of LLMs to generate a dataset of Neural Networks}

\author{Nadia Daoudi \\
Luxembourg Institute of Science and Technology, Luxembourg\\
nadia.daoudi@list.lu
\and
Jordi Cabot \\
Luxembourg Institute of Science and Technology, Luxembourg\\ 
University of Luxembourg, Luxembourg \\
jordi.cabot@list.lu}

\date{}
\maketitle

\begin{abstract}
Neural networks are increasingly used to support decision-making.
To verify their reliability and adaptability, researchers and practitioners have proposed a variety of tools and methods for tasks such as NN code verification, refactoring, and migration.
These tools play a crucial role in guaranteeing both the correctness and maintainability of neural network architectures, helping to prevent implementation errors, simplify model updates, and ensure that complex networks can be reliably extended and reused.
Yet, assessing their effectiveness remains challenging due to the lack of publicly diverse datasets of neural networks that would allow systematic evaluation.
To address this gap, we leverage large language models (LLMs) to automatically generate a dataset of neural networks that can serve as a benchmark for validation.
The dataset is designed to cover diverse architectural components and to handle multiple input data types and tasks.
In total, 608 samples are generated, each conforming to a set of precise design choices.
To further ensure their consistency, we validate the correctness of the generated networks using static analysis and symbolic tracing.
We make the dataset publicly available to support the community in advancing research on neural network reliability and adaptability.


\end{abstract}

\section{Background \& Summary}

Neural networks (NNs) have become an integral component of modern software systems, enabling them to perform complex prediction and decision-making tasks.
Their widespread adoption spans various domains and applications, and they are increasingly at the core of software development.
To support the design and development of NNs, a variety of tools have been proposed to ensure their reliability and adaptability.
Verification has become a key research area, ensuring that models behave as expected and satisfy critical properties of safety or robustness~\cite{leofante2018automated}. 
In parallel, refactoring techniques have emerged to enhance models efficiency and structure without severely altering their behaviour~\cite{wang2025refactoring}. 
Given the rapid evolution of NN technologies, several tools have been developed to facilitate NN migration either by transferring trained, ready-to-use models or by migrating the NN code itself across frameworks~\cite{liu2020enhancing,nn_migration}. 

Existing NN-related datasets are primarily designed for architecture benchmarking or to support NN training by providing samples for prediction tasks~\cite{jawahar2023llm,11080380}.
Consequently, evaluating techniques that support NN reliability and adaptability (e.g., migration, refactoring) remains challenging due to the lack of standardized benchmark datasets focused on diversity of NN architectures.
In practice, these approaches are often evaluated using a few NNs selected from previous studies, making it difficult to conduct thorough and fair assessment. 
To properly evaluate the robustness and generality of NN tools, a diverse dataset of architectures is needed.
However, manually collecting or designing such models is time-consuming and does not guarantee representativeness. 
This limitation slows progress and hinders fair and consistent evaluation across NN tools.

Large language models (LLMs) have revolutionized various domains, ranging from natural language understanding to software development and scientific computing~\cite{electronics14183580}.
Besides their traditional applications, LLMs have shown great potential in generating synthetic datasets.
For example, they have been used to generate tabular data samples to support NN training ~\cite{wang2024harmonic}.
Moreover, LLMs have shown promising results in code generation, producing functional code based on detailed user instructions~\cite{bistarelli2025usage}.
Given these capabilities, LLMs stand as a promising tool for creating diverse and high-quality code-related datasets that could help overcome existing limitations in evaluation and data availability.

In this paper, we propose a dataset of NN architectures generated using LLMs.
Specifically, we rely on GPT-5~\cite{openai2025gpt5} to create a diverse collection of NN models designed to support research on NN reliability and adaptability.
Carefully crafted instructions were used to guide the generation to ensure diversity in architecture and design characteristics.
The generated NNs can handle various input data types and perform a range of prediction tasks.
Moreover, they are designed with different complexity levels to better reflect the variability of design that generally exist in NN models.
In total, 608 NNs were generated, representing, to the best of our knowledge, the first dataset of NN architectures created to facilitate research on NN reliability and adaptability.
We also developed a tool that performs static analysis and symbolic tracing to ensure the validity of the architectures and that they adhere to the instructions in the prompt.
We make our dataset publicly available to support the community and encourage further research on NN development.

\section{Methods}
\begin{figure}
\centering
\includegraphics[width=0.9\textwidth]{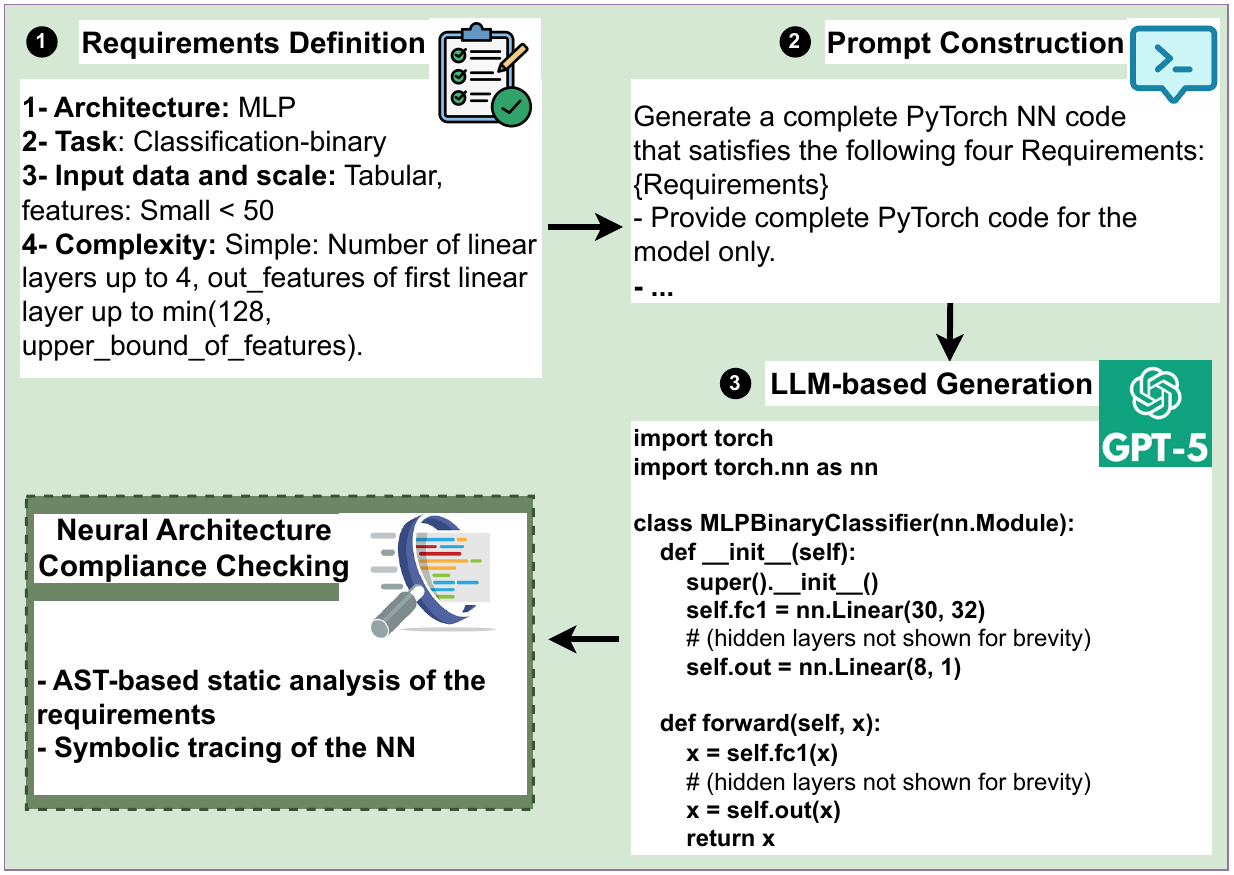}
\caption{Overview of our LLM-driven NN dataset generation}
\label{fig:overview}
\end{figure}

Our NN dataset was created in three main steps: Requirements Definition, Prompt Construction, and LLM-based Generation, which are illustrated in Figure~\ref{fig:overview}.

\subsection{Requirements Definition}\label{section:methodology:requirements}
The first step in the process is to define the characteristics of the NN architectures to be generated.
Since LLMs need specific instructions to guide the generation, it is important to feed them with precise requirements for the desired NN architectures.
Specifically, we define four categories of requirements to ensure diversity:

\subsubsection{Architecture}: It refers to the type of layers used in the NN definition.
Seven categories are considered: \textit{MLP}, \textit{CNN-1D}, \textit{CNN-2D}, \textit{CNN-3D}, \textit{RNN-Simple}, \textit{RNN-LSTM}, and \textit{RNN-GRU}.
These categories are chosen to guarantee that our dataset is diverse and covers the fundamental layers in NNs.
We note that each of these architecture types is characterised by the use of a specific NN layer, hereafter referred to as a \textit{characterising layer (CL)}, as presented in the first two columns of Table~\ref{table:architecture_layer_complexity}.

\begin{table}[!ht]
\centering
\caption{CL and complexity thresholds of NN architectures}\label{table:architecture_layer_complexity}
\scalebox{0.96}{
\begin{tabular}{c c | c c c}

\multicolumn{2}{c|}{\textbf{Architecture}} & \multicolumn{3}{c}{\textbf{Complexity}} \\
\hline
\textbf{Type} & \textbf{CL} & \textbf{\makecell[c]{Width \\ Thr}} & \textbf{\makecell[c]{Width \\ Variable}} & \textbf{\makecell[c]{Depth \\ Thr}} \\
\hline
MLP   & Linear     & $min(128, U)$ & $out\_features$ & 4 \\ 
CNN-1D & Conv1d  & $min(128, U)$   & $out\_channels$   & 4 \\ 
CNN-2D & Conv2d  & $min(8, W//8)$   & $out\_channels$   & 4 \\ 
CNN-3D & Conv3d  & $min(8, W//8)$   & $out\_channels$   & 4 \\ 
RNN-Simple   & RNN      & $min(128, U)$   & $hidden\_size$    & 2 \\ 
RNN-LSTM  & LSTM     & $min(128, U)$   & $hidden\_size$    & 2 \\ 
RNN-GRU   & GRU      & $min(128, U)$   & $hidden\_size$    & 2 \\ 
\hline 
\multicolumn{5}{l}{Thr: Threshold; $U: upper\_bound\_features$; $W: image\_width$}  
\end{tabular}
}
\end{table}

\subsubsection{Task}\label{section:methodology:requirements:task}
It refers to the prediction task performed by the NN.
We considered four main tasks to make the NNs diverse: \textit{Binary-classification}, \textit{Multiclass-classification}, \textit{Regression} and \textit{Representation-learning}.
We note that we excluded \textit{Multilabel-classification} as it closely resembles \textit{Binary-classification}: both use a sigmoid output (explicitly or via losses like \textit{BCEWithLogitsLoss}), differing mainly in the number of output neurons.
Hence, including it would not contribute to the diversity of the dataset.

\subsubsection{Input type and scale}
NNs can learn from a variety of input data types, each having unique characteristics and scales.
These differences affect the learning process and determine which architectural choices are most effective for capturing the underlying structure of the data.
Besides input type, we have also selected specific scales or ranges for input to drive the generation of NNs with diverse architectures.
Changes in the dimensionality of input data affects how layers are structured and connected, and can also influence the inclusion of components such as normalization or activation layers, resulting in NNs with varied layer composition.
We considered four input data types, \textit{Tabular}, \textit{Time Series}, \textit{Text}, and \textit{Image}, each with its corresponding input scale.
These types and scales are summarized in columns 1 and 3 of Table~\ref{table:possible_combinations}.

\subsubsection{Complexity}
It influences the number of layers and the width of NNs, and is used to guide the generation of networks ranging from simple to more advanced structures.
The complexity has been tailored to each architecture type, reflecting its distinct requirements for input data and intended tasks.
We define two thresholds: one for the number of \textit{CLs} and one for the width of the first CL.

Table~\ref{table:architecture_layer_complexity} summarizes these width and depth thresholds, along with the width variable, for the different architecture types.
In total, four complexity levels were considered: (1) \textit{Simple}, referring to an NN with a basic architecture, where the number of \textit{CLs} and the width of the first \textit{CL} (c.f., Table~\ref{table:architecture_layer_complexity}) are expected to be below the thresholds; (2) \textit{Wide}, an NN where the width of the first \textit{CL} is at least the width threshold; (3) \textit{Deep}, an NN whose total number of \textit{CLs} is at least the depth threshold; and (4) \textit{Wide-Deep}, an NN whose number of \textit{CLs} and width of the first \textit{CL} are at least the respective thresholds.
For example, in a \textit{simple MLP}, the number of \textit{Linear} layers does not exceed four and the first Linear layer contains up to \textit{min(128, upper\_bound\_of\_features)} output neurons.
Here, \textit{upper\_bound\_of\_features} is either 50 or 2000, as defined by \textit{Input type and scale} requirement of tabular data.

For \textit{CNN-2D} and \textit{CNN-3D}, the width of the first \textit{CL} is defined as $min(8, W//8)$, where $W$ is the image width. 
The value of 8 was chosen to ensure that \textit{simple} NNs remain relatively small even for large images, while \textit{wide} NNs still exceed this baseline for smaller images. 
For \textit{MLP}, \textit{CNN-1D}, and \textit{RNNs} (i.e., \textit{RNN-Simple}, \textit{RNN-LSTM} and \textit{RNN-GRU}) architectures, the first \textit{CL} width is defined as $min(128, U)$, with $U$ the number of input features.
128 was chosen as a round, sufficiently large value to allow \textit{wide} NNs to scale with the input while keeping \textit{simple} NNs relatively small. 
These specific values were selected to systematically preserve the relative distinction between \textit{simple} and \textit{wide} NNs across varying input sizes.

For \textit{Text} input type, \textit{CNN-1D} and \textit{RNNs} architectures can be used.
However, since the input is text, the network starts with an \textit{Embedding} layer rather than the standard \textit{CL} of the architecture.
The width threshold is therefore applied to this \textit{Embedding} layer.
Specifically, the width variable corresponds to \textit{embedding\_dim} of the \textit{Embedding} layer, with a threshold of $min(128, upper\_bound\_of\_vocab\_size)$.

To avoid ambiguity in complexity types, we add the precise definition to the complexity level based on both the architecture type and the input type and scale.
An example of the full complexity definition is shown in Step 1 of Figure~\ref{fig:overview}.

\subsection{Prompt Construction}\label{section:methodology:prompt}

{\captionsetup{type=listing}  
\caption{LLM prompt template}}
\begin{promptbox}[label={box:llm_prompt}]
Generate a complete PyTorch NN code that satisfies the following four Requirements: \\

Architecture: \textcolor{blue}{architecture}\par
Learning Task: \textcolor{blue}{task}\par
Input Type + Scale: \textcolor{blue}{input}\par
Complexity: \textcolor{blue}{complexity}\\

- Provide complete PyTorch code for the model only.\par
- All layer parameters must be fixed values inside the network, with no variables or external constants used.\par
- Avoid creating excessively large layers that would cause unreasonable memory usage.\par
- The model must only use standard layers directly, without defining helper functions or custom block classes.\par
- Output only the code, with no comments.
\end{promptbox}

Generating NN code with an LLM requires a carefully crafted prompt.
We base our instructions on the requirements outlined in the previous section, providing details about the network architecture, input type and scale, prediction task and complexity.
This approach ensures that the generated NN code follows the specified requirements, producing a variety of NNs that collectively cover a wide range of architectural characteristics and input data configurations.
We present in Listing~\ref{box:llm_prompt} our prompt template.

In the template, \textit{Architecture}, \textit{Learning Task}, \textit{Input Type + Scale}, and \textit{Complexity} are variables that change with each generation.
We also provide additional instructions to the LLM, such as ensuring that the generated code is fully self-contained, avoiding the use of parameters defined outside the network, which could otherwise introduce unnecessary dependencies.
We further avoid introducing custom block classes or helper functions, ensuring the generated networks are expressed in the standard way supported by the framework, keeping the dataset structured and consistent.

\begin{table}[!thb]
\centering
\caption{Combinations of NN requirements considered for dataset generation}\label{table:possible_combinations}
\scalebox{0.97}{
\begin{tabular}{c | c c c c}

 & \multicolumn{4}{c}{\textbf{Possible combinations}} \\
\hline
\textbf{\makecell[c]{Input \\ Type}} & \textbf{Arch} & \textbf{Input Scale} & \textbf{T x C} & \textbf{Total} \\
\hline
\textit{Tabular}    & \makecell[c]{MLP} & \makecell[c]{$f<50, f>2000$ \\ (2 scales)} & 4 x 4 & 32 \\ \hline

\textit{\makecell[c]{Time \\ series}} & \makecell[c]{CNN–1D, \\ RNN-Simple, \\ RNN-LSTM, \\ RNN-GRU }  &  \makecell[c]{$s<50 $  $\&$  $ f<50,$ \\ $s<50 $  $\&$  $ f>2000,$ \\ $s>2000$  $\&$  $f<50,$ \\ $s>2000 $  $\&$  $ f>2000$ \\ (4 scales)}  & 4 x 4 & 256 \\ \hline

\textit{Text}  & \makecell[c]{CNN–1D, \\ RNN-Simple, \\ RNN-LSTM, \\ RNN-GRU }  &  \makecell[c]{$s<50 $  $\&$  $ v<1k, $\\ $s<50 $  $\&$  $ v>100k,$ \\ $s>2000 $  $\&$  $  v<1k, $\\ $s>2000 $  $\&$  $ v>100k $\\ (4 scales)} & 4 x 4 & 256 \\ \hline
\textit{Image}  & \makecell[c]{CNN-2D, \\ CNN-3D} & \makecell[c]{$r<64, r>1024  $\\ (2 scales)}   & 4 x 4 & 64 \\ \hline
\multicolumn{5}{l}{Arch: Architecture; T: Task; C: Complexity } \\
\multicolumn{5}{l}{$f$: $features$; $s$: $seq\_length$; $v$: $vocab\_size$; $r$: $resolution$}

\end{tabular}}

\end{table}

\subsection{LLM-based Generation}\label{section:methodology:generation}
We employ \textit{GPT-5}~\cite{openai2025gpt5} to produce the NN code.
We use the requirements defined in Section~\ref{section:methodology:requirements} to populate the prompt template presented in Section~\ref{section:methodology:prompt}.
To that end, we first determine the possible combinations of requirements as some combinations are irrelevant or do not correspond to valid NN architectures. 
For example, using an \textit{MLP} to process \textit{image} inputs is not a typical design choice, as it does not leverage the spatial structure of the data.
We present a summary of the possible combinations of requirements in Table~\ref{table:possible_combinations}.
In total, our dataset contains 608 NNs generated by \textit{GPT-5}.

\section{Data Record}

A sample in our NN dataset is a Python file containing the NN model class definition.
Each file includes the prompt used to generate the code at the top.
File names encode four pieces of information in the format: \textit{architecture\_task\_input-type-scale\_complexity.py}.
For input scale, we avoid exact numbers and use \textit{small} or \textit{large} to indicate the scale of \textit{features}, \textit{sequence length}, \textit{vocabulary size}, or \textit{resolution}.
For example: \textit{mlp\_classification-binary\_tabular-large\_simple.py}.
The dataset is publicly available on GitHub\footnote{\url{https://github.com/BESSER-PEARL/LLM-Generated-NN-Dataset}}.

\section{Data Overview}

Diversity is a key aspect of our dataset.
While the NN requirements specified for the generation define certain NN characteristics, such as the architecture type, the exact layers and tensorops used to construct the NNs are purely determined by the LLM.
We therefore inspected the NN components in our dataset and found a total of 6842 layers and tensorops, comprising 38 unique types, which confirms the diversity of the structural components.

\begin{figure}
\centering
\includegraphics[width=\columnwidth]{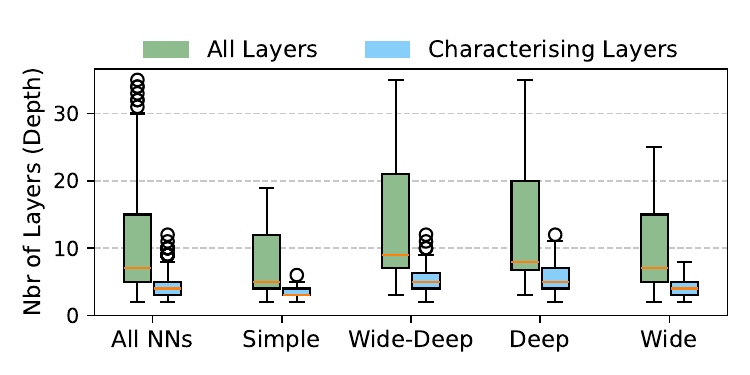}
\caption{Depth of all layers and of \textit{CLs} for the full NN dataset and for NNs grouped by complexity level}
\label{fig:depth_all}
\end{figure}

In the complexity requirements, we explicitly defined deep NNs as those with a number of \textit{CLs} above a certain threshold. 
While this threshold sets the minimum number of \textit{CLs}, the exact number of \textit{CLs}, as well as all other layers, is determined by the LLM.
We examine the distribution of NN depth in Figure~\ref{fig:depth_all}, which presents boxplots showing the depth of all layers and of \textit{CLs} for both the full dataset and for NNs grouped by complexity. 
The overall depth of NNs ranges from 2 to 35. 
For NNs grouped by complexity, the patterns are consistent: \textit{Simple} NNs generally have smaller depth (both for all layers and \textit{CLs}), followed by \textit{Wide}, \textit{Deep}, and \textit{Wide-Deep} NNs. 
For \textit{Wide} NNs, no depth threshold was set; only the width was constrained. 
Nevertheless, the LLM produced slightly deeper NNs compared to \textit{Simple} NNs, which could be explained by the fact that \textit{Wide} NNs require more layers to perform feature transformations.
Overall, the NN dataset contains a variety of layers and tensorops and includes NNs of different depths.

\section{Technical Validation}

This section describes the tool developed to ensure that NN architectures satisfy the prompt specifications.
The tool verifies the compliance with the four requirements described in Section~\ref{section:methodology:requirements}. 
In addition, it validates that the architectures are structurally correct and execute without errors.
Our tool uses static analysis to inspect the NN architectures.
This is achieved using \textit{AST}~\cite{ast} that parses the NN code to extract all layers, including functional calls~\cite{pytorch_functional}, and their execution order.
Additionally, symbolic tracing is employed to verify the structural integrity of the NN architectures.
Our validation tool is available in our dataset repository.

The tool verifies that the NN contains at least one of its \textit{CLs} to confirm the architecture. 
Regarding the task requirement, it ensures that the output layer configuration matches the expected task, with the appropriate number of output neurons for tasks such as \textit{classification} and \textit{regression}.
For the input type and scale, the tool inspects the first \textit{CL} to verify that they match the expected specification. 
This includes checking parameters such as \textit{in\_features} for \textit{tabular} data, and \textit{in\_channels} or \textit{input\_size} for \textit{time series}. 
Symbolic tracing is integrated at this stage to ensure that the input is compatible with the NN structure and scale, using values derived from the expected input sizes.
For complexity, the tool verifies NN depth and width by comparing the number of \textit{CLs} and \textit{CL} parameters with the thresholds specified in Table~\ref{table:architecture_layer_complexity}. 

We ran our tool on the NNs to verify their compliance with the requirements.
In total, 8 out of 608 NNs were flagged by our tool as non-compliant, all belonging to the \textit{RNN-LSTM} and \textit{RNN-GRU} categories and have large \textit{time series} input features.
These NNs share a common issue: a \textit{Linear} projection layer was used before the \textit{CL} (\textit{LSTM} or \textit{GRU}), which results in breaking the sequential structure of the input data and preventing the temporal modelling.
To maintain consistency, the 8 NNs were regenerated, and their compliance was successfully verified by our tool.

We note that we further selected, for each input data type, one NN and trained it on standard benchmark datasets~\cite{krizhevsky2009learning,NIPS2015_250cf8b5,pace1997sparse,anguita2013public}.
The results showed consistent performance, confirming the correctness and usability of the code.

\section{Related Work}
In this section, we review related works and describe the originality of our NN dataset compared to existing datasets.
\subsection{Neural Network Datasets}
Existing datasets related to NNs generally focus on Neural Architecture Search (NAS)~\cite{JMLR:v20:18-598}, which aims to automate the process of NN design to achieve the highest performance.
NAS-Bench-101~\cite{pmlr-v97-ying19a} has proposed a set of CNN architectures mapped to their performance metrics.
Similarly, NAS-HPO-Bench~\cite{klein2019tabular} and NAS-HPO-Bench-II~\cite{pmlr-v157-hirose21a} have considered the problem of joint optimization.
Recently, Younger~\cite{yang2024younger} has been proposed as a dataset containing NN architectures represented as directed acyclic graphs to assist in NN design and optimization.
Other traditional datasets have been also widely used to train NNs for a range of different tasks~\cite{GONG2023107268,ma15041428,hamid2018benchmark,krizhevsky2009learning,russakovsky2015imagenet,netzer2011reading}.

Our work differs from these approaches by providing a dataset of diverse NNs designed to meet specific requirements on architecture, task, input, and complexity. 
This diversity makes it suitable for NN reliability and adaptability methods such as NN verification~\cite{leofante2018automated,huang2017safety}, refactoring~\cite{wang2025refactoring,kaplunovich2020refactoring}, and migration~\cite{liu2020enhancing,nn_migration,besser-nn}. 
In contrast, the above approaches focus on NN training or architecture search, which would prioritise performance and limit the diversity of the NNs.

\subsection{LLM-based Dataset Generation}

LLMs have been used for NAS in several works~\cite{jawahar2023llm,NEURIPS2023_184c1e18,10.1145/3638529.3654017,song2024position}.
Besides, they have been leveraged for post-architecture design tasks, such as generating NN parameters~\cite{soroinstruction,wang2025neurogen}.
Synthetic data generation for NN training has also thrived thanks to LLMs that can create diverse samples across different domains~\cite{11080380,10.1007/978-3-031-93418-6_9,long2024llms,ronval2025tagal}.
While the main focus of these approaches is performance optimization, our work uses LLMs to generate a diverse dataset of NN architectures that can be leveraged by NN verification and migration methods.

\section{Data Availability}
Our NN dataset is publicly available in our repository\footnote{\url{https://github.com/BESSER-PEARL/LLM-Generated-NN-Dataset}}.

\section{Code Availability}
The code used to generate our dataset is publicly available in our repository\footnote{\url{https://github.com/BESSER-PEARL/LLM-Generated-NN-Dataset}}.

\section{Author Contribution}
The authors of the manuscript have contributed equally to this work.

\section{Competing interests}
The authors declare no competing interests.

\section{Funding}
This project is supported by the Luxembourg National Research Fund (FNR) PEARL program, grant agreement 16544475.


\end{document}